\documentclass[smallcondensed]{svjour3}     
\smartqed  

\usepackage{booktabs}
\usepackage{xcolor}
\usepackage{lmodern}
\usepackage[misc]{ifsym}
\usepackage{graphicx}
\usepackage{amssymb}
\usepackage{lineno}
\modulolinenumbers[5]
\usepackage{booktabs}
\usepackage{multirow}
\usepackage{hyperref}
\usepackage{amsmath}

\usepackage[ruled,lined]{algorithm2e}

\begin{document}

\title{Exceedance Probability Forecasting via Regression: A Case Study of Significant Wave Height Prediction
}

\titlerunning{Exceedance Probability Forecasting via Regression}        

\author{Vitor~Cerqueira
        \and
        Luis~Torgo
}

\authorrunning{V. Cerqueira et al.} 

\institute{V. Cerqueira (\Letter) \at
         Faculdade de Engenharia da Universidade do Porto, Porto, Portugal\\
         LIACC, Porto, Portugal\\
         \email{vcerqueira@fe.up.pt}
         \and
         Lu\'{i}s Torgo \at
         Dalhousie University, Halifax, Canada\\
}

\date{Received: date / Accepted: date}

\maketitle

\begin{abstract}

Significant wave height forecasting is a key problem in ocean data analytics. This task affects several maritime operations, such as managing the passage of vessels or estimating the energy production from waves.
In this work, we focus on the prediction of extreme values of significant wave height that can cause coastal disasters.
This task is framed as an exceedance probability forecasting problem. Accordingly, we aim to estimate the probability that the significant wave height will exceed a predefined critical threshold.
This problem is usually solved using a probabilistic binary classification model or an ensemble of forecasts. Instead, we propose a novel approach based on point forecasting. Computing both type of forecasts (binary probabilities and point forecasts) can be useful for decision-makers. While a probabilistic binary forecast streamlines information for end-users concerning exceedance events, the point forecasts can provide additional insights into the upcoming future dynamics.
The proposed solution works by assuming that the point forecasts follow a distribution with the location parameter equal to that forecast. Then, we convert these point forecasts into exceedance probability estimates using the cumulative distribution function. 
We carried out experiments using data from a smart buoy placed on the coast of Halifax, Canada. The results suggest that the proposed methodology is better than state-of-the-art approaches for exceedance probability forecasting.

\end{abstract}

\section{Introduction}\label{intro}

Significant wave height (SWH) forecasting is a key problem in ocean data analytics, and several methods have been developed for tackling it \cite{huang1998empirical,duan2016hybrid,ali2020near}. Forecasting the ocean conditions is valuable for multiple operations, such as vessel performance. A decision support system based on these forecasts can optimize ship performance in operations such as selecting the best routes, speed, or heading \cite{guo2023evaluating}.
Moreover, the passage of vessels requires a minimum depth of water for their movement. The occurrence of extreme values of SWH reduces the depth of water and this minimum may not be met. 

Another motivation for forecasting SWH is related to renewable energy, where forecasts are used to estimate energy production.  Predicting impending large values of SWH is important for coastal disaster prevention and protection of wave energy converters, which should be shut down to prevent their damage \cite{li2012wave}. 
In summary, accurate SWH predictions can increase the performance of vessels, reduce costs, and improve the reliability of ports.

In this work, we focus on the prediction of large values of SWH.
We frame this problem as an exceedance probability forecasting task. Exceedance probability forecasting denotes the process of estimating the probability that a time series will exceed a predefined threshold in a predefined future period. This task is usually relevant in domains where extreme values (i.e., the tail of the distribution) are highly relevant, such as earthquakes, and hurricanes \cite{kunreuther2002risk,frangopol2014prognosis}.

\begin{figure}[h]
    \centering
    \includegraphics[width=\textwidth, trim=0cm 0cm 0cm 0cm, clip=TRUE]{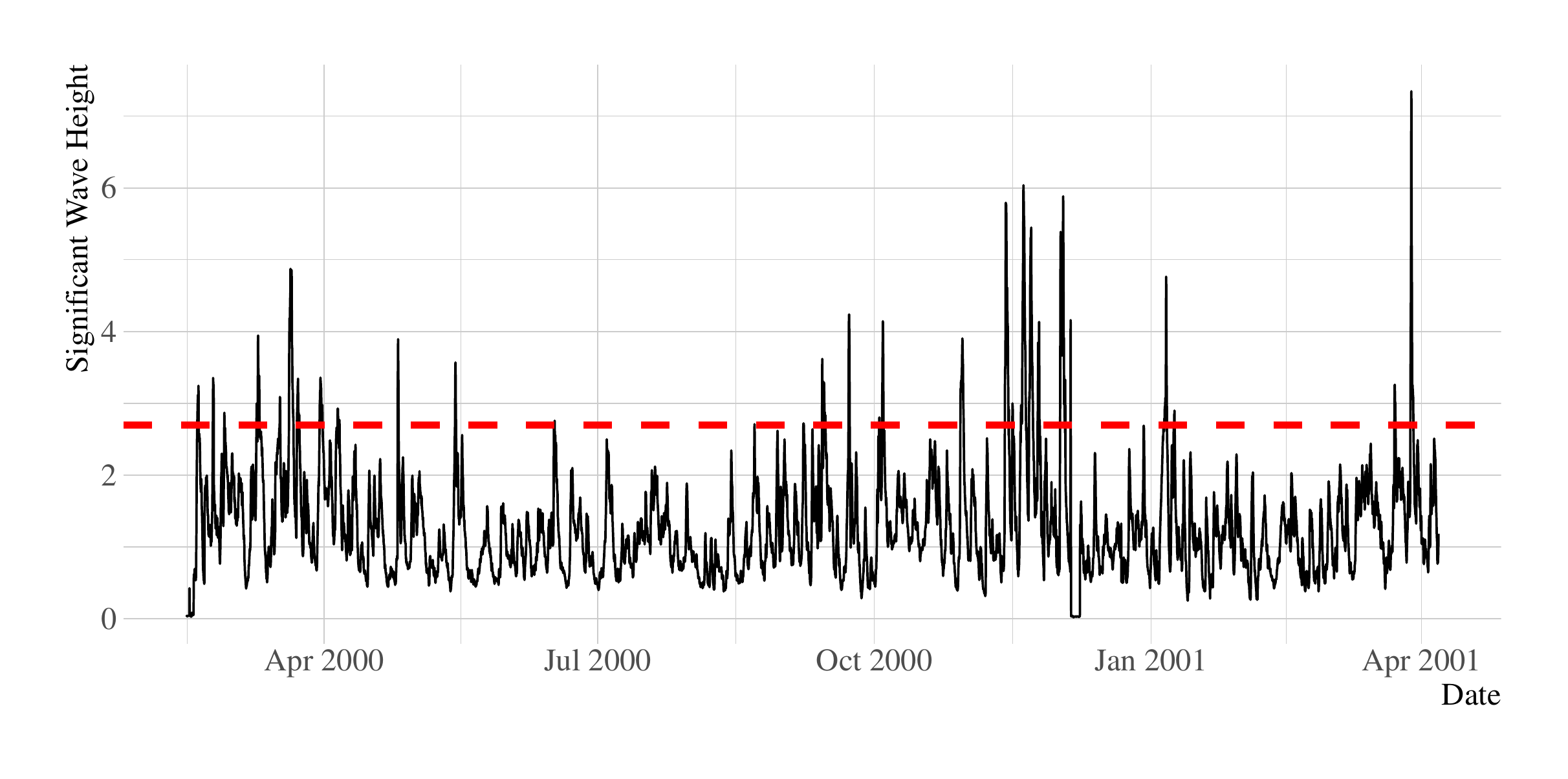}
    \caption{Significant wave height time series with hourly granularity. The red horizontal dashed line represents the exceeding threshold.}
    \label{fig:ts_subset}
\end{figure}

Figure \ref{fig:ts_subset} shows an hourly SWH time series. The red horizontal dashed line represents the threshold above which a specified exceedance event occurs. The objective is to predict these events early in time.

Exceedance forecasting is typically formalized as a binary classification problem, where the target variable denotes whether or not the threshold will be exceeded. Ensemble-based forecasting (regression) approaches can also be applied, in which the exceedance probability is estimated according to the ratio of individual predictions that exceed the threshold. Both approaches often model these events using lagged features, i.e., an auto-regressive type of modelling \cite{taylor2017probabilistic}.

In this paper, we propose a novel approach to exceedance probabilistic forecasting. The method is based on the numeric point forecasts, for example, produced by an auto-regressive model. The proposed method does not rely on an ensemble-based forecasting model, though it could be one. 
Essentially, we assume that a given forecast is modeled using a continuous distribution. In this case, we resort to a Weibull distribution following previous works on modeling SWH data \cite{muraleedharan2007modified}. Then, we use the cumulative distribution function (CDF) of this distribution to estimate the exceedance probability. 
The point forecast represents the location parameter of the distribution, while the remaining parameters (e.g. scale) are estimated using the training data.

Using a forecasting model instead of a classifier to obtain exceedance probability estimates is useful because of the additional information. Binary probabilities concerning exceedance events streamline the information being given to the end-user. Coupling this information with the numeric forecasts for upcoming observations offers extra information. This extra information might be valuable for understanding the data dynamics. For example, when the data approaches a threshold value, the projected trajectory based on point forecasts offers valuable insights into the upcoming changes in the variable.

We apply the methodology in a case study related to SWH forecasting.
The results suggest that coupling a forecasting model with the proposed mechanism based on the CDF is better for exceedance probability forecasting relative to a classification method or an ensemble-based approach.

In summary, the contributions of this work are the following:
\begin{itemize}
    \item A novel approach for estimating time series exceedance probability based on point forecasting and the CDF;
    
    \item The application of this method to estimate the exceedance probability of SWH, which is important for managing maritime operations. 
\end{itemize}

The code for the methods used in the paper is publicly available\footnote{\url{https://github.com/vcerqueira/exceedance_wave}}.
The rest of the paper is organized as follows. In Section \ref{sec:rw}, we overview the related work. In Section \ref{sec:pd}, we define the predictive problem, followed by Section \ref{sec:method}, where the proposed method is formalized. The case study used in this work is described in Section \ref{sec:cs}. The experiments carried out to validate the proposed method are presented in Section \ref{sec:experiments}. We discuss the results and conclude the paper in Section \ref{sec:discussion}.

\section{Related Work}\label{sec:rw}

This section provides an overview of previous works related to this paper. 
We start by reviewing the literature on SHW forecasting (Section \ref{sec:rw1}). Then, we describe approaches for probabilistic forecasting focusing on exceedance probability estimation (Section \ref{sec:rw2}).

\subsection{Significant Wave Height Forecasting}\label{sec:rw1}

Ali et al. \cite{ali2020near} analyze buoy data from eastern Australia. They apply several regression approaches, including a model tree and multivariate adaptive regression splines, to forecast short-term (30 minutes in advance) values of SWH. They conclude that a hybrid model, optimized using a method based on least squares, provided the best performance.

Duan et al. \cite{duan2016hybrid} model to forecast SWH time series several hours in advance. Unlike a short-term forecast, a horizon of a few hours is important to ensure the safety of maritime operations, e.g. passage of vessels. They apply the empirical mode decomposition (\cite{huang1998empirical}) to decompose the time series into several simpler components, which are then modeled using an auto-regressive approach. This decomposition leads to better forecasting performance relative to an auto-regressive approach without this process.

Shamshirband et al. \cite{shamshirband2020prediction} compared different regression algorithms for significant wave height forecasting, including artificial neural networks, support vector regression, and extreme learning machines \cite{huang2006extreme}. They concluded that all methods perform comparably, though the extreme learning machine model showed a slight advantage.

Several authors attempt to apply different neural network architectures for SWH predictions, including vanilla recurrent neural networks \cite{sadeghifar2017coastal}, feedforward neural networks \cite{londhe2006one,jain2007real}, gated recurrent units \cite{wang2021forecasting}, LSTMs \cite{wang2021forecasting,abdullah2022significant,bethel4153300assessing}, or graph neural networks \cite{chen2021significant}. A comparison of these architectures is not available. However, there is a tendency for recurrent architectures, which is expected due to their capabilities for modeling sequential data. In our work, we use a deep feedforward neural network, among other regression methods. We found that, in our case study, this architecture performed better than a recurrent one. Some architectures based on a multi-layer perceptron architecture show a competitive forecasting performance with transformers and recurrent-based architectures \cite{challu2023nhits}.

Dixit et al. \cite{dixit2015removing} also model SWH using a neural network. However, instead of using an auto-regressive type of modeling, they apply a discrete wavelet transform as a preprocessing step. They claim that this step helps remove prediction lag, thereby improving forecasting performance for SWH multistep ahead forecasting.

Fasuyi et al. \cite{fasuyi2020machine} and Kuppili \cite{kuppili2021forecasting} also model smart buoy data to interpolate data across different buoys located in distinct locations. When a buoy is shut down for maintenance, the data from other nearby buoys can be used to interpolate the missing values of that buoy.

The works mentioned so far are similar to each other in the sense that all attempt to forecast the future values or interpolate from near buoys, of SWH time series using machine learning regression models. It is difficult to compare the performance of models across different works due to the usage of different data and performance estimation methods. However, larger forecasting horizons lead to poorer forecasting performance, as expected.


\subsection{Exceedance Probability Forecasting}\label{sec:rw2}

Conveying the uncertainty around forecasts is critical for better decision-making. In standard time series forecasting tasks, uncertainty quantification is usually carried out with prediction intervals \cite{khosravi2011comprehensive}, or probabilistic forecasting \cite{gneiting2014probabilistic}.
When modeling binary events, it is common to tacke this problem using a classification approach \cite{taylor2016using}. An instance of a binary event is exceedance, which, as mentioned before, refers to when a time series exceeds a predefined threshold.

Slud et al. \cite{slud1994partial} apply logistic regression to model the exceedance probability of rainfall. In their model, the lagged observations of the time series are used as predictor variables in an auto-regressive fashion.

Taylor et al. \cite{taylor2016using} use an exceedance probability approach for financial risk management. They predict whether the returns of financial assets will exceed a threshold. This is accomplished by using a method called CARL (conditional auto-regressive logit), which is based on logistic regression. Later, CARL was extended to handle problems involving multiple thresholds by  Taylor et al. \cite{taylor2017probabilistic}. The extended model was used to predict wind ramp events, where the tails on both sides of the distribution are important to predict.

The alternative to using a classification approach to estimate the exceedance probability is to use an ensemble of forecasts. Accordingly, the probability is computed as the ratio of forecasts above the threshold. This type of approach is commonly used in fields such as meteorology \cite{lewis2005roots}, where different forecasts are obtained by considering different initial conditions \cite{ferro2007comparing}. Toth et al. \cite{toth2003probability} provide a comprehensive description of ensemble-based approaches listing their main properties.
Estimating the exceedance probability for different thresholds enables the construction of exceedance probability curves \cite{kunreuther2002risk}, which may be helpful for better decision-making.

Exceedance probability forecasting is related to extreme value prediction. The goal of forecasting extreme values is to accurately predict the numeric value of extreme observations in time series. For example, Polson et al. \cite{polson2020deep} forecast energy load spikes using a deep learning approach. On the other hand, in exceedance probability, the goal is simplified to predict whether or not a given threshold is exceeded, i.e., the magnitude of exceedance is not a major concern.

\section{Problem Definition}\label{sec:pd}

This section formalizes the predictive task addressed in this work. We start by defining SWH forecasting problems (Section \ref{sec:tsf}), followed by the formalization of exceedance probability forecasting (Section \ref{sec:exc}).

\subsection{Time Series Forecasting}\label{sec:tsf}

A univariate time series represents a temporal sequence of values $Y = \{y_1, y_2, \dots,$ $y_n \}$, where $y_i \in \mathcal{Y} \subset \mathbb{R}$ is the value of $Y$ at time $i$ and $n$ is the length of $Y$. 
The general goal behind forecasting is to predict the value of the upcoming observations of the time series, $y_{n+1}, \ldots, y_{n+h}$, given the past observations, where $h$ denotes the forecasting horizon. 

We formalize the problem of time series forecasting based on an auto-regressive strategy. Accordingly, observations of a time series are modeled based on their recent lags. 
We construct a set of observations of the form ($X$, $y$) using time delay embedding according to the Takens theorem~\cite{Takens1981}. 
In each observation, the value of $y_{i}$ is modelled based on the most recent $q$ known values: $X_i = \{y_{i-1}, y_{i-2}, \dots, y_{i-q} \}$, where $y_{i} \in \mathcal{Y} \subset \mathbb{R}$,
which represents the observation we want to predict, $X_i \in \mathcal{X} \subset \mathbb{R}^q$ represents the $i$-th embedding vector. 
For each horizon, the goal is to build a regression model $f$ that can be written as $y_{i} = f(X_i, Z_i)$, where $Z_i$ represents additional covariates known at the $i$-th instance. In effect, we address the multi-step ahead forecasting problem with a direct method \cite{taieb2012review}. We build $h$ models, one for predicting each future step $\{1, \dots, h\}$. 
The case study in this work is based on a multivariate time series that is described in Section \ref{sec:data_set}. In this case,
the covariates $Z$ are explanatory time series related to SWH, including wind speed or sea surface temperature. Each of the explanatory time series is also represented at each time step $i$ with an embedding vector based on the past recent values. This leads to an auto-regressive distributed lags modeling technique \cite{nkoro2016autoregressive}.

\subsection{Exceedance Probability Forecasting}\label{sec:exc}

The probability of exceedance $p_i$ denotes the probability that the value of a time series will exceed a predefined threshold $\tau$ in a given instant $i$.
Typically, $p_{i}$ is modelled by resorting to a binary target variable $b_{i}$, which can be formalized as follows:
\begin{equation}\label{eq:task1b}
  b_{i} = \begin{cases}
            1 & \text{if } y_{i} \geq \tau ,\\
            0 & \text{otherwise}.
          \end{cases}
\end{equation}

\noindent Essentially $b_{i}$ takes the value of 1 if the threshold $\tau$ is exceeded is a future period $i$, or 0 otherwise. This naturally leads to a classification problem where the goal is to build a model $g$ of the form $b_{i} = g(X_i, Z_i)$. Accordingly, $p_{i}$ represents the conditional expectation of $b_{i}$ which we estimate based on Bernoulli density. Logistic regression is commonly used to this effect \cite{taylor2016using}.

Since the exceedance events are rare, the number of cases for the positive class (i.e. when $b_{i} = 1$) is significantly smaller than the number of cases for the negative class ($b_{i} = 0$). This leads to an imbalanced classification problem, which hinders the learning process of algorithms \cite{branco2016survey}.

The alternative to using a classification model is to use a forecasting ensemble $F$ composed of a set of $M$ models: $F = \{f_1, f_2, \dots, f_M\}$. These models can be combined for forecasting using the uniformly weighted average as follows:
$$
\hat{y}_{i} = \frac{1}{M}\sum_{k=1}^M f_k (X_i,Z_i)
$$

We can also use the ensemble $F$ to estimate the exceedance probability according to the ratio of ensemble members that forecast a value that exceeds the threshold $\tau$:
$$
p_{i} = \frac{1}{M}\sum_{k=1}^M I(f_k(X_i,Z_i) \geq \tau)
$$

\noindent where $I$ denotes the indicator function. Hereafter, we will refer to this approach as an ensemble-based direct method for exceedance probability. Ensemble-based approaches are commonly used in some fields related to environmental science, for example, meteorology or hydrology \cite{toth2003probability}.

\section{Methodology}\label{sec:method}

The goal of this paper is to tackle exceedance probability forecasting problems using a case study concerning SWH prediction.
Since large values of SWH are important to anticipate, we design an approach to estimate the probability of their occurrence. To our knowledge, this is the first attempt at framing SWH forecasting as an exceedance probability estimation task. 
In this section, we describe the proposed approach for estimating the exceedance probability (Section \ref{sec:method1}). We also list the main advantages of the method relative to state-of-the-art approaches (Section \ref{sec:method2}).

\subsection{Exceedance Probability via the CDF}\label{sec:method1}

The predictive task is to estimate $p$, the probability that the SWH will exceed a specified threshold $\tau$. We tackle this task using a forecasting model that produces point forecasts. The point forecasts, in addition to the exceedance probability estimates, can provide valuable information about the dynamics of the SWH. In effect, building a model capable of both types of forecasts is desirable for operators.

This work introduces a novel approach for exceedance probability forecasting based on a time series forecasting model. 
The proposed solution works in two main steps:
\begin{itemize}
    \item Building a forecasting model and use it to obtain point forecasts
    \item Converting point forecasts into exceedance probability estimates using the CDF
\end{itemize}

\subsubsection{Building a forecasting model}

We build a forecasting model $f$ using auto-regression as formalized in Section \ref{sec:tsf}. The value of future observations are modeled based on their past lags.
However, it is important to remark that the proposed methodology is agnostic to the underlying forecasting model.

The proposed method leverages the numeric predictions produced by the forecasting model regarding the value of future observations. Here, we denote the point forecast for the $i$-th observation as $\hat{y}_i$.

\subsubsection{Computing exceedance probability estimates}

We assume that $\hat{y}_i$ can be modeled according to a right-skewed Weibull distribution with a location parameter equal to the forecast $\hat{y}_i$. The scale $\beta$ and shape $\alpha$ parameters are estimated using the training data. We use the Weibull distribution as it provides an adequate fit to SWH data \cite{muraleedharan2007modified}. For other case studies, a different distribution may be more appropriate.
In these conditions, we can estimate $p_i$, the exceedance probability at the $i$-th time step, using the corresponding CDF. This is done as follows:

\begin{equation}
    p_i = 1 - F(\tau; \hat{y}_i, \beta, \alpha)
\end{equation}

\noindent Where $F$ denotes the CDF of the Weibull distribution.
When evaluated at the threshold $\tau$, the CDF represents the probability that the respective random variable will take a value less than or equal to $\tau$. In effect, we subtract $F(\tau; \hat{y}_i, \beta, \alpha)$ from $1$ to obtain the probability that the random variable will exceed $\tau$. This process is described in algorithm \ref{alg:mc}.

\IncMargin{1em}
\begin{algorithm}[ht]
    \SetKwInOut{Input}{Input}
    \SetKwInOut{secInput}{    }
    \SetKwInOut{Output}{Output}

    \Input{$f$: Forecasting model}
    \secInput{$\tau$: Exceedance threshold}
    \secInput{$y_{train}$: Training data}
    \Output{$p_{i}$}

    \BlankLine

    $\beta, \alpha \leftarrow  $ $weibull.fit(y_{train})$ \tcp{Fit the Weibull distribution using the training data and get its parameters}
    
    $\hat{y}_i \leftarrow $ $f$($\{y_{i-1}, y_{i-2}, \dots, y_{i-q} \}$)  
    \tcp{Compute the forecast for the $i$-th time-step}

    $p_i \leftarrow $ $1 - F(\tau; \hat{y}_i, \beta, \alpha)$\tcp{Estimate the exceedance probability according to the CDF}
    
    Return $p_i$
\caption{Computing the exceedance probability estimate using point forecasts}
\label{alg:mc}
\end{algorithm}
\DecMargin{1em}


\subsection{Methodological Advantages}\label{sec:method2}

Aside from performance considerations, which we will delve into in Section \ref{sec:experiments}, the proposed method may be preferable relative to a classifier for two main reasons. The first reason is integration because the same forecasting model can be used to predict both the upcoming values of SWH and to estimate exceedance probability. 
The exceedance probability estimate simplifies the information being given to the end user, which may help a more inexperienced operator in their decision-making. Yet, complementary numeric forecast offers more information, which might be valuable for more experienced operators. For example, if the forecasts are near the threshold value, the predicted trajectory offers insights into how the variable is expected to change.

The second reason is threshold flexibility: a classification approach fixes the threshold during training, and it cannot be changed during inference. Conversely, a regression model follows a lazy approach regarding the threshold. Therefore, in a given instant we can use the same model to estimate the exceedance probability for different thresholds -- this enables the plotting of exceedance probability curves.
These advantages apply to an ensemble-based forecasting approach. However, the high computational costs and lack of transparency often preclude the application of these strategies \cite{lakkaraju2016interpretable,cerqueira2021model}.

\section{Case Study}\label{sec:cs}

In this section, we present the case study we use in the experiments. First, we describe the dataset and the respective source (Section \ref{sec:data_set}). Then, we explain the preprocessing steps carried out before applying the predictive models to the time series (Section \ref{sec:data_prep}).

\subsection{Data Set}\label{sec:data_set}

We conducted our study using data collected from the Environment and Climate Change Canada Buoy (ECCC Buoy). This buoy is located 13 kilometers away in the open sea close to Halifax, Canada. The data from this buoy was previously used for an interpolation task by Fasuyi et al. \cite{fasuyi2020machine}, as we described in Section \ref{sec:rw1}.

The time series spans from 11-02-2000 15:00:00 to 01-04-2020 11:00:00 and includes the following variables:
\begin{itemize}
    \item SWH: Significant wave height (in meters) which represents the variable we want to predict;
    
    \item PWP: Peak wave period (in seconds), which is the wave period carrying the maximum amount of energy;
    
    \item CWH: Maximum zero crossing wave height (in meters);
    
    \item WDIR: Wind direction (in degrees);
    
    \item WSPD: Horizontal wind speed (in meters per second);
    
    \item GSPD: Gust wind speed (in meters per second);
    
    \item DRYT: Buoy dry bulb temperature (in degrees Celsius);
    
    \item SSTP: Approximated sea surface temperature (in degree Celsius).
\end{itemize}

\begin{figure}[h]
    \centering
    \includegraphics[width=\textwidth, trim=0cm 0cm 0cm 0cm, clip=TRUE]{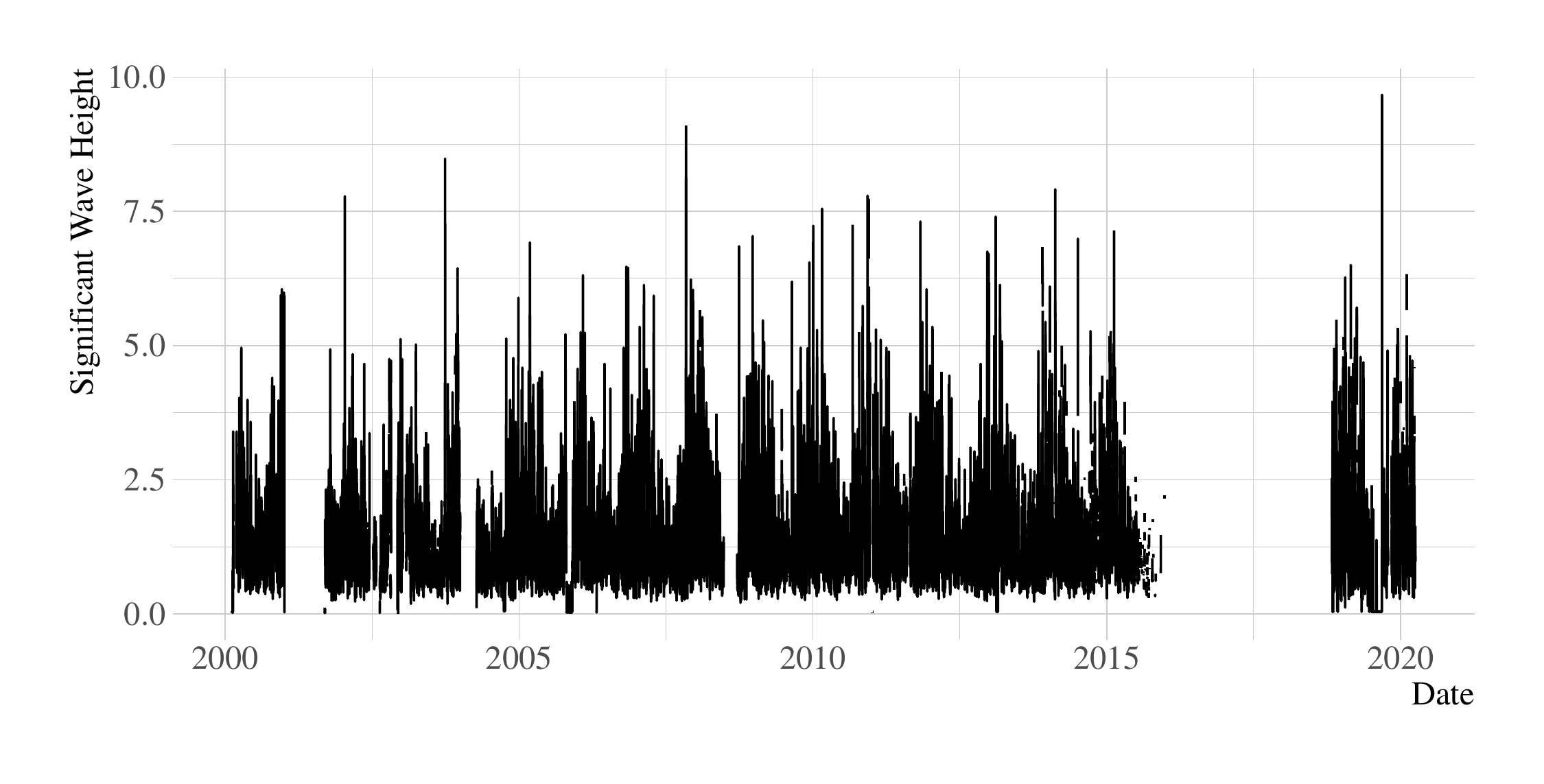}
    \caption{SWH time series collected with the ECCC buoy.}
    \label{fig:tseries}
\end{figure}

\noindent All variables of the time series are regularly sampled at an hourly frequency. 
Figure \ref{fig:tseries} shows the SWH time series. For visualization purposes, we only plot the response variable (SWH).
The total number of observations is 125.890. However, there is a considerable number of missing observations, especially in the period from February 2016 to October 2019.

\begin{figure}[h]
    \centering
    \includegraphics[width=\textwidth, trim=0cm 0cm 0cm 0cm, clip=TRUE]{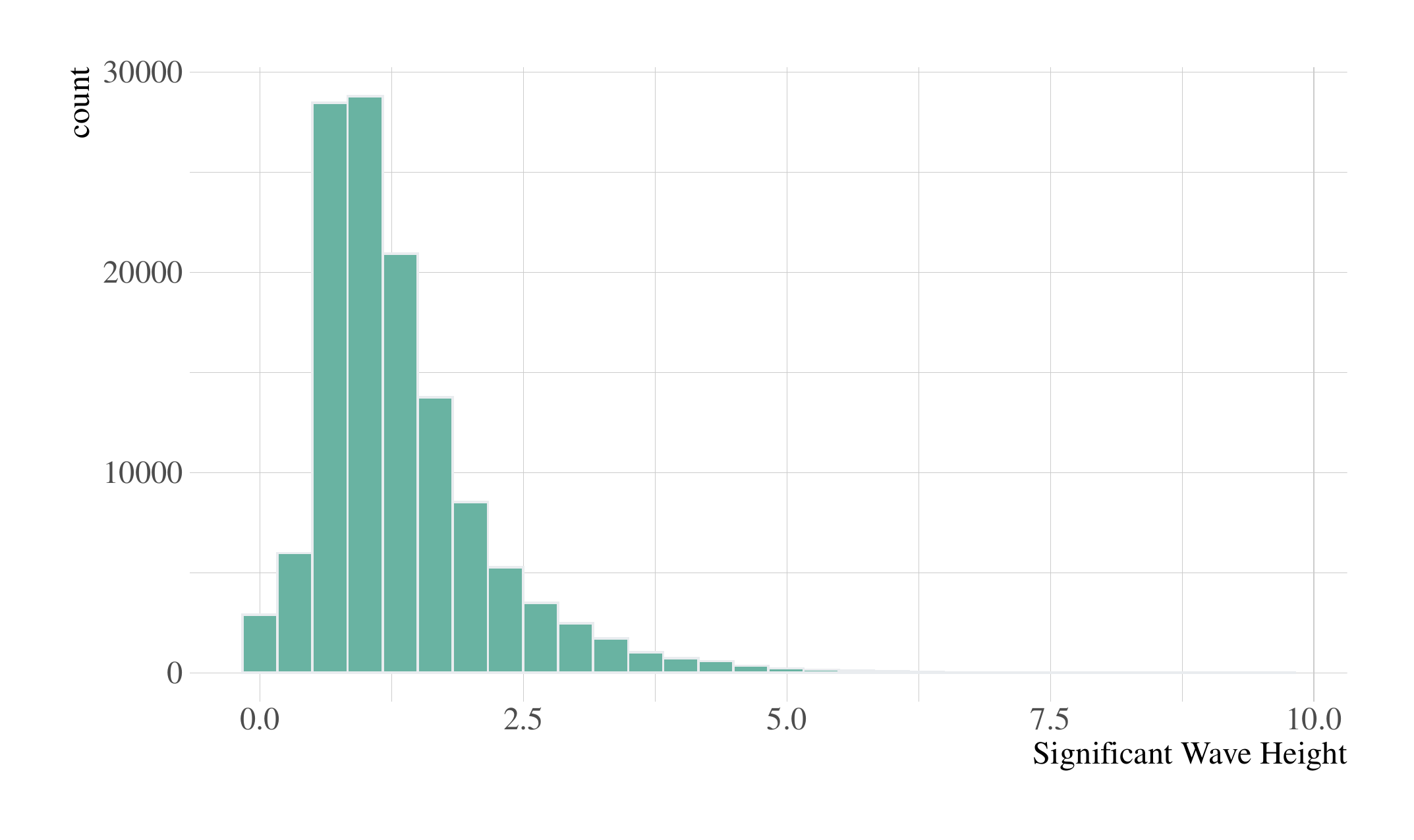}
    \caption{Histogram of the SWH time series distribution, which shows a positive skewness}
    \label{fig:tseriesh}
\end{figure}

In Figure \ref{fig:tseriesh}, we show a histogram that illustrates the distribution of the SWH time series. The distribution is right-skewed where the heavy tail represents the large SWH sea state we attempt to predict.

\begin{figure}[h]
    \centering
    \includegraphics[width=.8\textwidth, trim=0cm 0cm 0cm 0cm, clip=TRUE]{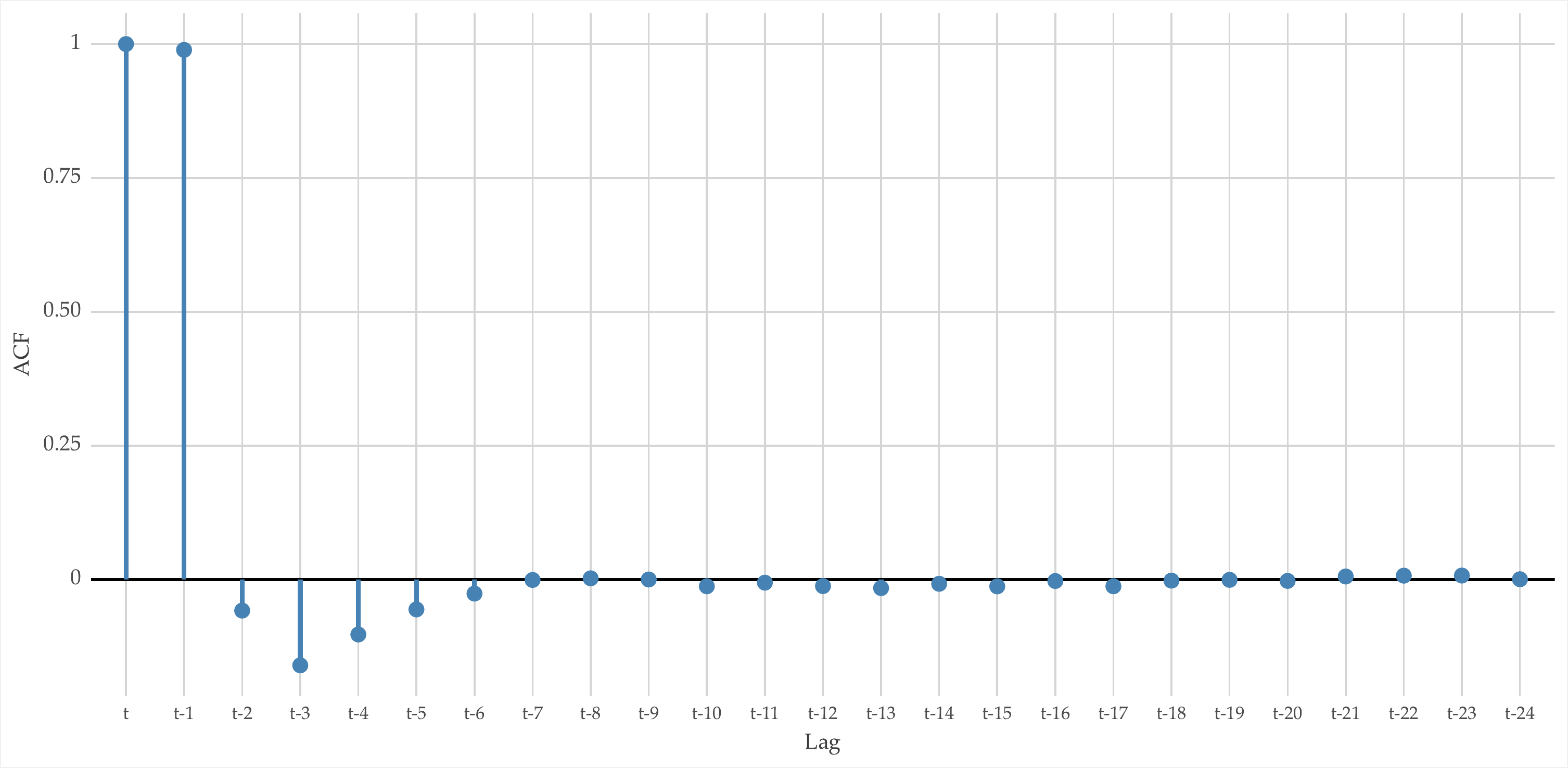}
    \caption{Partial auto-correlation structure of the SWH time series.}
    \label{fig:acf_scores}
\end{figure}

Figure \ref{fig:acf_scores} shows the partial auto-correlation scores of the SWH time series up to 24 lags. The result shows that the variations in SWH can be explained with the first few lags.

\subsection{Data Preparation}\label{sec:data_prep}

\begin{figure}[h]
    \centering
    \includegraphics[width=.85\textwidth, trim=0cm 0cm 0cm 0cm, clip=TRUE]{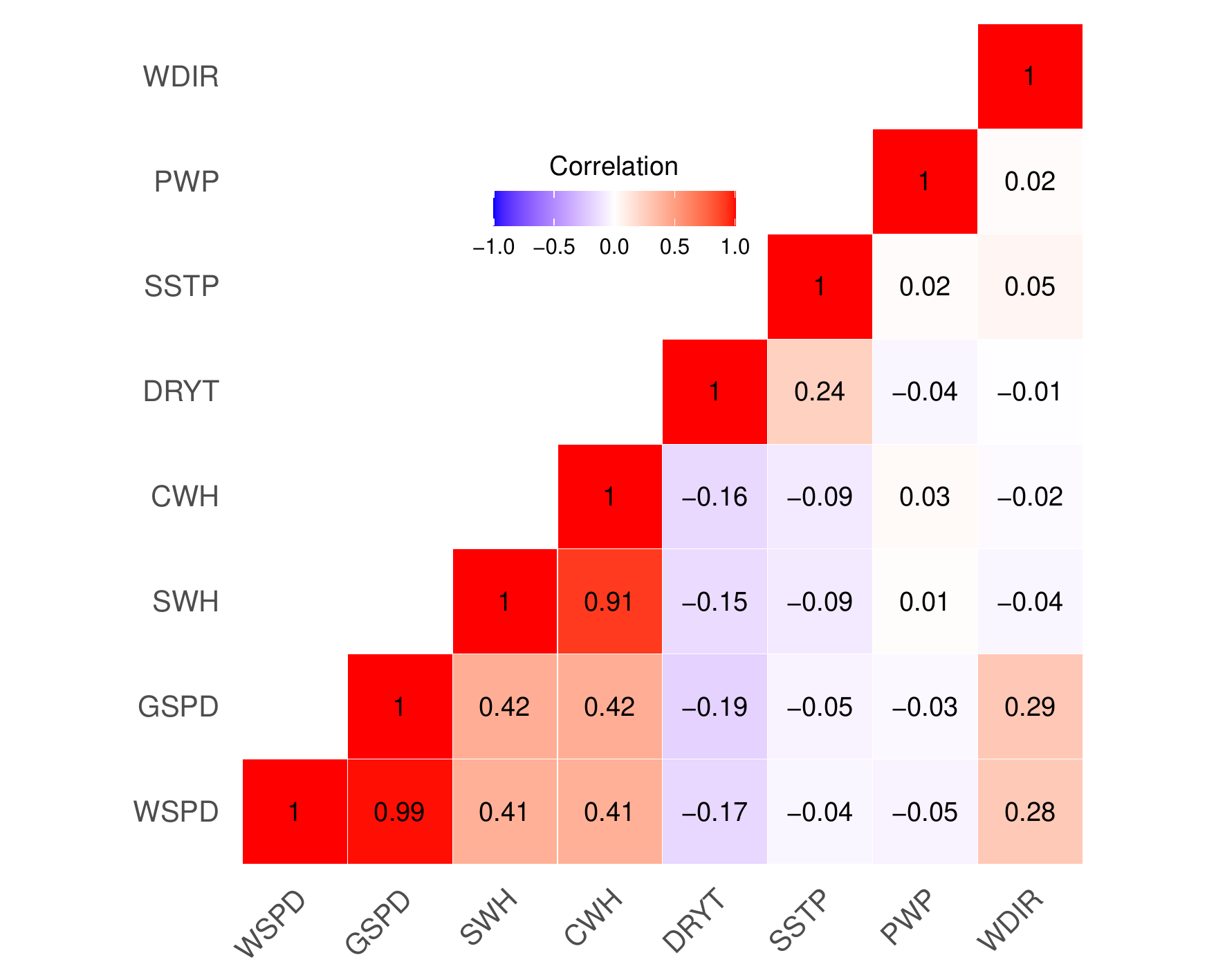}
    \caption{Heatmap showing the pairwise correlation across all variables in the data set.}
    \label{fig:corrmat}
\end{figure}

As exploratory data analysis, in Figure \ref{fig:corrmat} we show the Pearson correlation of each pair of variables in the data set. Overall, the main variable SWH is highly correlated with CHW (0.91), and moderately correlated with the wind and gust speed variables (WSPD and GSPD) with values 0.41 and 0.42, respectively. 

We constructed a set of observations for training the predictive models according to the formalization in Section \ref{sec:tsf}. The target variable $y$ represents the value of SWH observed $h$ steps (hours) ahead in the future. The respective binary target variable $b$ is computed according to whether $y$ takes a value above $\tau$ in $h$ hours.
In order to measure the predictability of exceeding events in different time frames, we analyzed multiple forecasting horizons. Specifically, we experiment with a horizon from 1 to 24 hours in advance. For example, with $h=6$ we attempt to predict whether the threshold will be exceeded in 6 hours.

In terms of the threshold $\tau$, we adopt a data-driven approach to set this value. Specifically, we set the threshold to be the 99\textit{th} percentile of the available SWH time series. We follow this approach, as opposed to a judgemental one (e.g. based on some domain expert advice), because it can be generalized to different data sets or domains.
In effect, at each iteration of the cross-validation procedure (which is detailed in the next section), we compute $\tau$ using the training data.
The average threshold across all iterations was $3.17$ meters.

Regarding the predictor variables, we follow the procedure described in Section \ref{sec:pd}.
We apply time delay embedding to all variables in the time series with an embedding size (number of lags) of 6. This means that, at a given instant, $y$ (or $b$) is modeled based on auto-regression using the past 6 values of each of the eight variables described above. We also added the day of the year in the explanatory variables to account for yearly seasonal effects. 
This leads to a total of 49 explanatory variables. 

We also performed a data validation step. 
The objective is to predict exceedance events. Therefore, if exceedance is occurring in the explanatory variables ($X$) it means that the event is already happening and prediction is not necessary at that moment. In this context, we remove the instances where exceedance is occurring in the explanatory variables because the model will not be used in this case.

\section{Experiments}\label{sec:experiments}

This section describes the experiments carried out to validate the proposed solution. These experiments cover the following research questions:

\begin{enumerate}
    \item \textbf{RQ1}: Which approach is better for SWH forecasting?;
    \item \textbf{RQ2}: How does the proposed method based on the CDF perform for estimating the exceedance probability relative to state-of-the-art approaches?;
    \item \textbf{RQ3}: What is the impact of the forecasting horizon on the results obtained?
\end{enumerate}

\subsection{Experimental Design}

We carry out a Monte Carlo cross-validation procedure to evaluate the performance of models \cite{picard1984cross}. This estimation method, which is also referred to as repeated holdout, has been shown to provide competitive forecasting performance estimates with univariate time series \cite{cerqueira2020evaluating}.
Monte Carlo cross-validation is applied with 10 folds. The training and test sizes in each fold are 50\% and 20\% of the size of the available data, respectively. 
In each fold, a point is randomly selected from the available data (constrained by the size of training and testing data). This point then represents the end of the training set, and the start of the testing set \cite{cerqueira2020evaluating}.

We evaluate the exceedance probability performance of each approach using the area under the ROC curve (AUC) and log loss as evaluation metrics. Besides, we also use typical regression metrics to measure forecasting performance, i.e., the performance for predicting the numeric value of future observations. Specifically, we use the mean absolute error (MAE), mean absolute percentage error (MAPE), coefficient of determination ($R^2$), and root mean squared error (RMSE).

\subsection{Methods}

In this subsection, we overview the approaches used to estimate the exceedance probability. These can be broadly split into classification methods (Section \ref{sec:clf}) and forecasting (regression) methods (Section \ref{sec:regr}).

\subsubsection{Classification Methods}\label{sec:clf}

We apply the following three classification methods for estimating exceedance probability:
\begin{itemize}
    \item Random Forest Classifier (\texttt{RFC}): A standard binary classifier built using a Random Forest \cite{breiman2001random};
    
    \item Logistic Regression (\texttt{LR}): Another standard classifier but created using the Logistic Regression algorithm. We specifically test this method because it is a popular one for exceedance probability tasks (c.f. Section \ref{sec:rw2});
    
    \item \texttt{RFC+SMOTE}: Exceedance typically involves imbalanced problems because the exceeding events are rare. We created a variant of \texttt{RFC}, which attempts to cope with this issue using a resampling method. Resampling methods are popular approaches for dealing with the class imbalance problem. In effect, we test a classifier that is fit after the training data is pre-processed with the SMOTE resampling method by Chawla et al. \cite{chawla2002smote};
\end{itemize}

\subsubsection{Forecasting Methods}\label{sec:regr}

We test four different regression approaches.
Namely, a Random Forest regression method (\texttt{RFR}), a \texttt{LASSO} linear model, and a deep feed-forward neural network (\texttt{NN}). These are three widely used regression algorithms for forecasting. 
The fourth approach is a heterogeneous regression ensemble (\texttt{HRE}). An ensemble is referred to as heterogeneous, as opposed to homogeneous, if it is comprised of base models trained with different learning algorithms. The following learning algorithms were used to train the base models of \texttt{HRE}: a rule-based model based on the \texttt{Cubist} algorithm, a bagging ensemble of decision trees (\texttt{Bagging}), ridge regression (\texttt{Ridge}), elastic-net (\texttt{ElasticNet}), \texttt{LASSO}, multivariate adaptive regression splines (\texttt{MARS}), light gradient boosting optimized using random search (\texttt{LGBM}); extra trees (\texttt{ExtraTree}), Adaboost  (\texttt{AdaBoost}), Random Forests (\texttt{RFR}), partial least squares regression (\texttt{PLS}), principal components regression (\texttt{PCR}), nearest neighbors (\texttt{KNN}), and a deep feedforward neural network (\texttt{NN}).

We applied different parameter settings and the total number of initial models is 50. The complete list of parameters is shown in Table \ref{tab:expertsspecs}. Using a validation set we trimmed the ensemble by removing half of the models with worse performance. The trimming process has been shown to improve the forecasting performance of ensembles \cite{jose2008simple,cerqueira2019arbitrage}. Usually, a dynamic combination rule improves the performance of forecasting ensembles relative to a static one. Notwithstanding, the simple average is a solid benchmark according to Cerqueira et al. \cite{cerqueira2019arbitrage}.

The neural network architecture contains 2 hidden layers with 32 and 16 units, respectively. Dropout was also included after each hidden layer with a drop rate of 0.2. All units followed a rectified linear unit activation function. Training occurred during 30 epochs. We optimized mean squared error using validation data and the Adam optimizer. 

\begin{table}[!bth]
	\centering
	\caption{Summary of the learning algorithms}		
	\begin{tabular}{llll}
	\toprule
	\textbf{ID} & \textbf{Algorithm} & \textbf{Parameter} & \textbf{Value}\\
	\midrule
	    \multirow{2}{*}{\texttt{MARS}} & \multirow{2}{*}{Multivar. A. R. Splines} & Degree & \{1, 3\} \\
	    
	    & & No. terms & \{10, 20\} \\
	    
	    & & Threshold & \{0.01, 0.001\} \\
	    
	    \midrule   
        
        \multirow{2}{*}{\texttt{LASSO}} & \multirow{2}{*}{LASSO Regression} & \multirow{2}{*}{L1 alpha} & \{0.25, 0.5, \\
        
	    & & & 0.75, 1\}\\
        
        \midrule   
        
        \multirow{2}{*}{\texttt{Ridge}} & \multirow{2}{*}{Ridge Regression} & \multirow{2}{*}{L2 alpha} & \{0.25, 0.5, \\
        
	    & & & 0.75, 1\}\\
        
        \midrule   
        
        \texttt{ElasticNet} & ElasticNet Regression &  \textit{Default} & - \\
        
	    \midrule   
	    
	    \multirow{3}{*}{\texttt{KNN}} & \multirow{3}{*}{k-Nearest Neighbors} & No. of Neighbors & \{1, 5, 10\} \\
	    
	    & & \multirow{2}{*}{Weights} & \{Uniform,\\
	    & & & Distance\}\\
	    
	    \midrule   
	    
	    \multirow{2}{*}{\texttt{RFR}} & \multirow{2}{*}{Random Forest Regression} & No. trees & \{50, 100\} \\
	    
	    & & Max depth & \{\textit{default}, 3\} \\
	    
        \midrule   
        
        \multirow{2}{*}{\texttt{ExtraTree}} & \multirow{2}{*}{Extra Trees Regression} & No. trees & \{50, 100\} \\
	    
	    & & Max depth & \{\textit{default}, 3\} \\
	    
	    \midrule   
	    
	    \multirow{2}{*}{\texttt{Bagging}} & \multirow{2}{*}{Bagging Regression} & \multirow{2}{*}{Base estimator} & \{Decision Stump, \\
        
	    & & & Decision Tree\}\\
	    
	    & & No. estimators & \{50, 100\} \\
	    
	    \midrule   
        
        \texttt{RBR} & Rule-based Regression & No. iterations & \{1, 5\}\\
        
        \midrule   
        
        \texttt{NN} & Deep Feedforward Neural Network & No. hidden layers & \{2\}\\
        
        \midrule   
        
        \multirow{2}{*}{\texttt{LGBM}} & \multirow{2}{*}{LightGBM} & \multirow{2}{*}{Booster} & \{DART, \\
        
	    & & & GOSS, GBDT\}\\
        
        \midrule   
        
        \multirow{3}{*}{\texttt{AdaBoost}} & \multirow{3}{*}{AdaBoost Regression} & \multirow{2}{*}{Base estimator} & \{Decision Stump, \\
        
	    & & & Decision Tree\}\\
	    
	    & & Learning rate & \{0.3, 0.7\} \\
	    
	    \midrule   
        
        \texttt{PCR} & Principal Comp. Regr. &  No. of components & \{2, 5\} \\
        
        \midrule   
        
        \texttt{PLS} & Partial Least Regr. & No. of components & \{2, 5\} \\
        
        \midrule  
        
        \texttt{OMP} & Orthogonal Matching Pursuit & \textit{Default} & - \\

		\bottomrule    
	\end{tabular}%
	\label{tab:expertsspecs}
\end{table}

We apply the regression approaches in two different ways to obtain the estimates for exceedance probability. The first is a \texttt{Direct} (abbreviated as \texttt{D}) approach: the exceedance probability $p$ is computed according to the ratio of ensemble members which predicts a value above the threshold $\tau$. The \texttt{Direct} approach is only valid for ensembles, in this case, \texttt{HRE} (a heterogeneous ensemble) and \texttt{RFR} (a homogeneous ensemble).
The second approach is using the CDF (\texttt{CDF}) according to our proposed method described in Section \ref{sec:method}. We denote which approach is used by appending the name to the regression model. For example, \texttt{HRE+CDF} represents the heterogeneous ensemble applied with the CDF for estimating exceedance probability.

\subsection{Results}

\begin{figure}[h]
    \centering
    \includegraphics[width=.9\textwidth, trim=0cm 0cm 0cm 0cm, clip=TRUE]{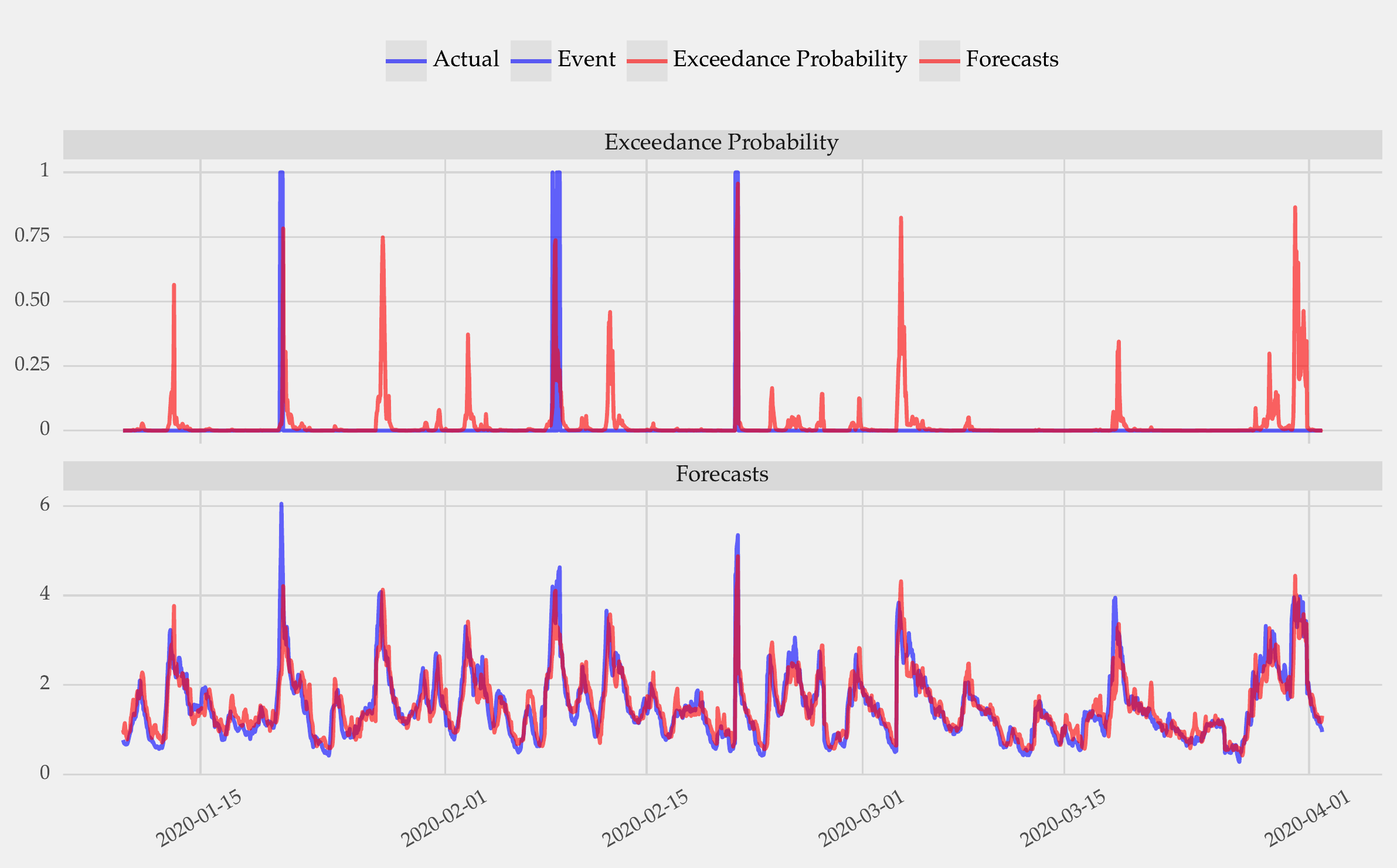}
    \caption{Sample of the forecasts and exceedance probabilities produced by the model.}
    \label{fig:exampleforecast}
\end{figure}

As a preliminary analysis, Figure \ref{fig:exampleforecast} shows a sample of the forecasts and exceedance probability estimates produced by the method \textbf{NN+CDF} for a lead time of 6 hours. The model can capture different exceedance events during this period.

\begin{figure}[h]
    \centering
    \includegraphics[width=\textwidth, trim=0cm 0cm 0cm 0cm, clip=TRUE]{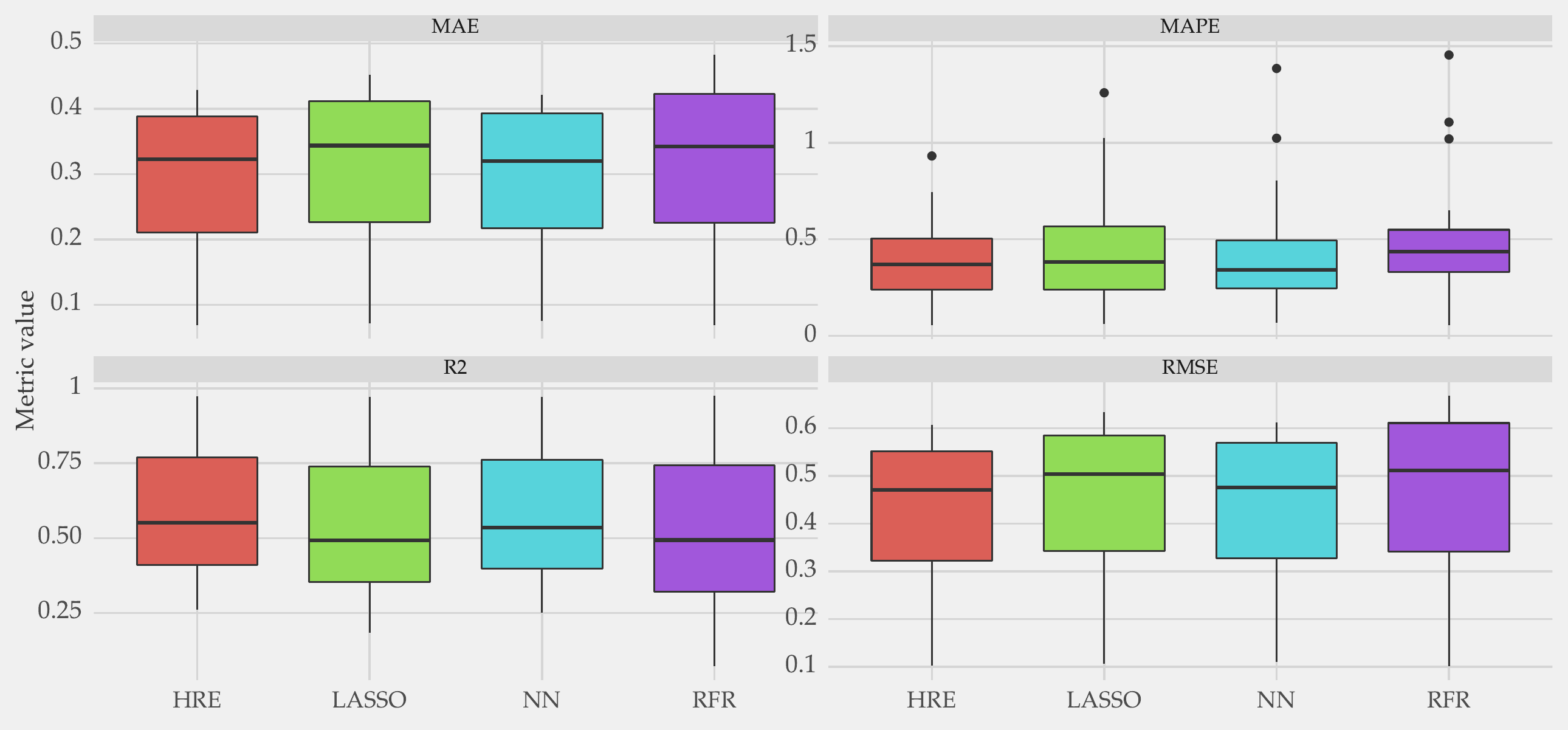}
    \caption{Distribution of forecasting performance across the forecasting horizon for each regression approach by different metrics.}
    \label{fig:r2_scores}
\end{figure}

In this subsection, we analyze the results of the experiments.
First, we analyze the performance of each regression method (\texttt{RFR}, \texttt{LASSO}, \texttt{HRE}, and \texttt{NN}) for forecasting the future values of the time series (\textbf{RQ1}). 
The results are shown in Figure \ref{fig:r2_scores}, which illustrates the distribution of performance of each method across the forecasting horizon with four metrics: MAE, MAPE, $R^2$, and RMSE. For all metrics, \texttt{NN} and \texttt{HRE} show the best performance, which is comparable with each other. These methods are followed by \texttt{RFR}.
The linear model \texttt{LASSO} underperforms relative to the others.

\begin{figure}[h]
    \centering
    \includegraphics[width=\textwidth, trim=0cm 0cm 0cm 0cm, clip=TRUE]{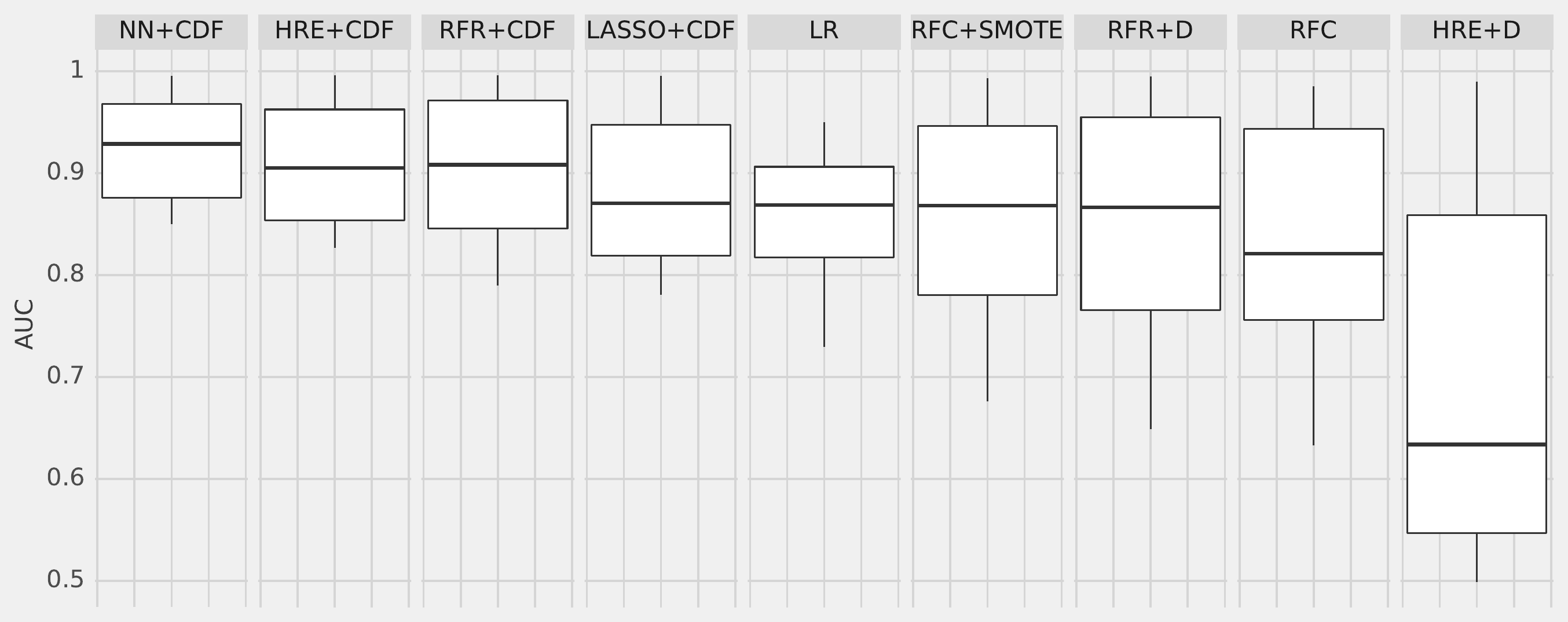}
    \caption{Distribution of the AUC for each method across the forecasting horizons.}
    \label{fig:auc_dist}
\end{figure}

We then evaluated all methods according to their ability to estimate exceedance probability. 
Figure \ref{fig:auc_dist} shows the distribution of AUC for each method across the forecasting horizons (up to 24 hours). 
The best performing models are \texttt{NN+CDF}, \texttt{RFR+CDF} and \texttt{HRE+CDF}, respectively. This rank is consistent with how these models performed for forecasting based on the regression-based scores. The results indicate that coupling a forecasting model with the mechanism based on the CDF for estimating exceedance probability provides a better performance relative to binary classification strategies (\texttt{RFC}, \texttt{RFC+SMOTE}, and \texttt{LR}) and ensemble-based approaches (\texttt{RFR+D} and \texttt{HRE+D}).

Another interesting result to note is that \texttt{HRE+D} provides poor estimates regarding exceedance probability, while \texttt{RFR+D} is competitive with classification methods.
This outcome suggests that building an ensemble with weak models (\texttt{RFR} is an ensemble of decision trees) is better than an ensemble of strong models, such as \texttt{HRE}. 

Pre-processing the training data with a resampling method (SMOTE) improved the average AUC of the \texttt{RFC}, though the difference is small. 
The linear model \texttt{LR} is comparable with the Random Forests classifiers.
On the other hand, \texttt{LASSO+CDF} underperforms relative to the other methods, except for \texttt{HRE+D}. Overall, these results answer the research question \textbf{RQ2}, regarding the competitiveness of the proposed approach in terms of performance.

\begin{figure}[h]
    \centering
    \includegraphics[width=\textwidth, trim=0cm 0cm 0cm 0cm, clip=TRUE]{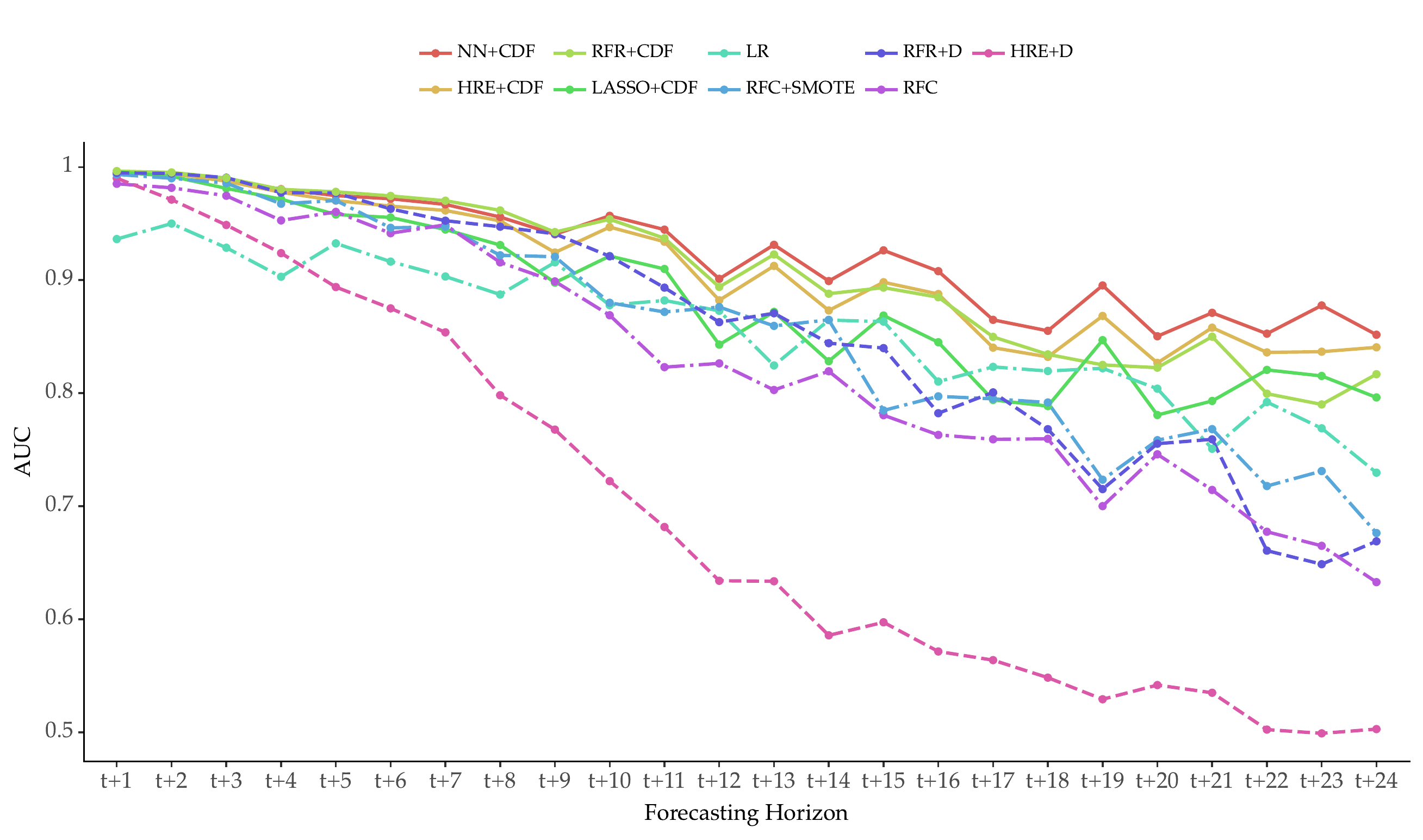}
    \caption{AUC for each method across the forecasting horizon. Classification methods (\texttt{RFC+SMOTE}, \texttt{RFC}, and \texttt{LR}) are represented with long dashed, ensemble direct methods (\texttt{RFR+D}, and \texttt{HRE+D}) are represented as dashed lines and \texttt{CDF}-based methods are denoted with solid lines.}
    \label{fig:roc_curve_over_h}
\end{figure}

Figure \ref{fig:roc_curve_over_h} shows the AUC score for each method across the forecasting horizons. There is a generalized tendency for decreasing scores as the forecasting horizon increases (\textbf{RQ3}). This is expected as predicting further into the future represents a more difficult task.

\begin{figure}[h]
    \centering
    \includegraphics[width=.9\textwidth, trim=0cm 0cm 0cm 0cm, clip=TRUE]{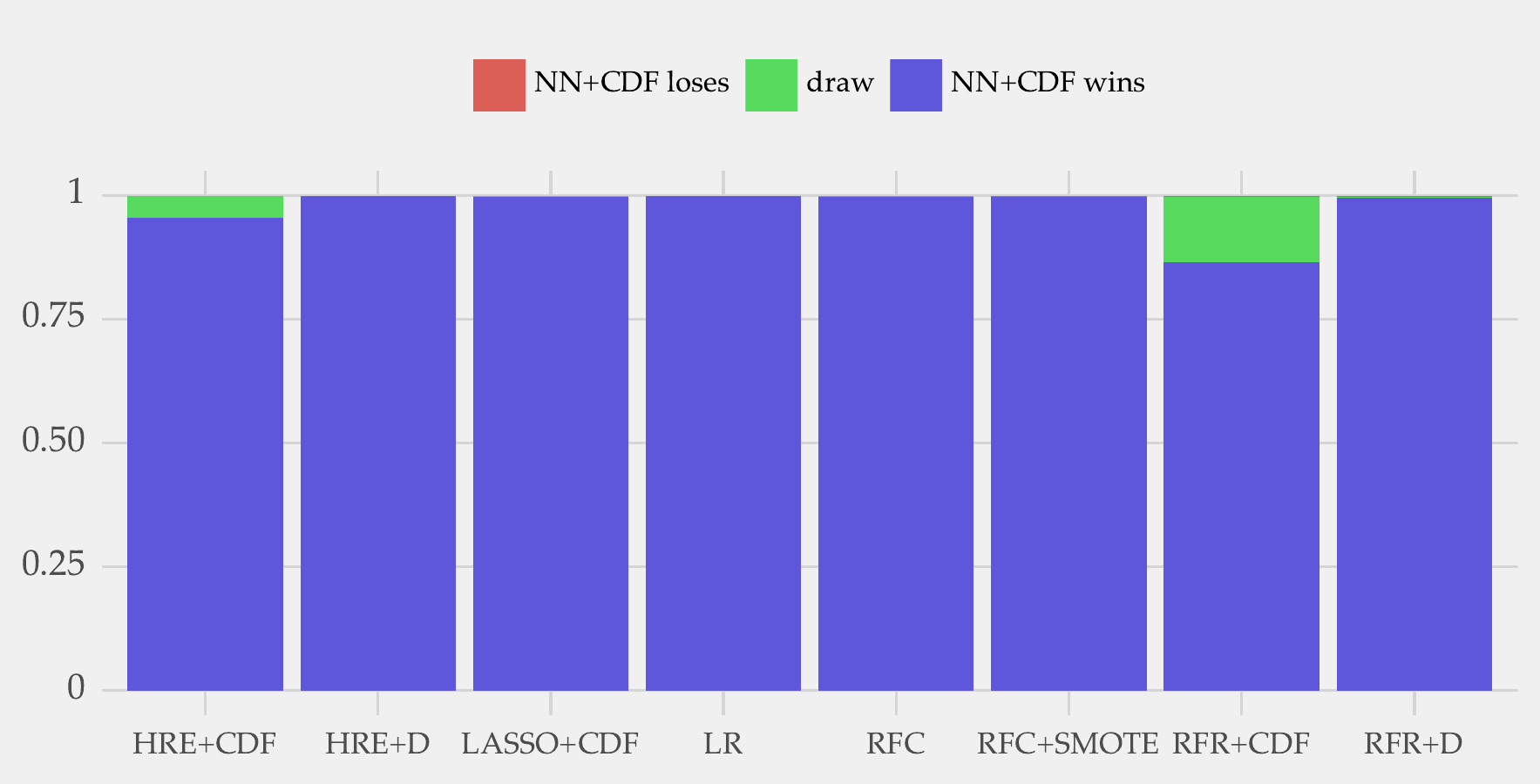}
    \caption{Results from the Bayesian correlated t-test. Probability of \texttt{NN+CDF} losing, drawing, or winning significantly against the respective method when considering a ROPE of 1\%.}
    \label{fig:bayesian}
\end{figure}

We carried out a Bayesian analysis to assess the significance of the results using the Bayesian correlated t-test \cite{benavoli2017time}. This test is used to compare pairs of predictive models across the forecasting horizon. In this case, we compare \texttt{NN+CDF} with all remaining methods.
We define the region of practical equivalence (ROPE) for the Bayes correlated t-test to be the interval [-1\%, 1\%]. This means that the performance of the two methods under comparison is considered equivalent if their percentage difference is within this interval. The results are presented in Figure \ref{fig:bayesian}, which shows the probability of \texttt{NN+CDF} winning in blue, drawing in green (results within the ROPE), or losing in red against each remaining method. \texttt{NN+CDF} outperforms other approaches, except for \texttt{HRE+CDF}. Notwithstanding, the ensemble \texttt{HRE+CDF} also follows the proposed approach for exceedance probability forecasting.

\section{Discussion}\label{sec:discussion}

The main conclusion from the experiments is that the proposed method provides better exceedance probability estimates relative to state-of-the-art alternatives for a case study concerning SWH forecasting. This outcome has an important practical impact because users (e.g. decision-makers or analysts responsible for maritime operations) can rely on a forecasting model to predict not only the value of future observations but also quantify the probability of an exceedance event (in this case, a large SWH state). 

A second important conclusion is that a deep learning neural network provided the best results when coupled with the proposed approach. Besides, a heterogeneous regression ensemble (\texttt{HRE}), whose members are trained with several different learning algorithms, did not provide better exceedance probability estimations than a Random Forest (\texttt{RFR}).

While exceedance probability forecasting usually refers to binary events (whether or not the exceedance occurs), multiple thresholds can also be defined.
An example of this is the triple barrier method used in quantitative trading, e.g. \cite{de2020machine}. Stock market traders may be interested in predicting buy, sell, or hold signals according to price movements. Exceeding a positive threshold in terms of predicted price returns can be used as a buy signal; the opposite (predicted price returns below a negative threshold) can represent a sell trigger. If none of the thresholds is met, then the trader should hold the current position.

We transformed the time series into a trainable format using time delay embedding on each variable, as we formalized in Section \ref{sec:pd}. 
During our experiments, we also tested a differencing operation to stabilize the mean, but this process did not impact performance.
The data shows a non-constant variance over time where the dispersion is higher in the winter months (c.f. Figure \ref{fig:tseries}). We tested the application of a Box-Cox method or a log transformation to stabilize the variance. However, this process also did not improve performance. A possible explanation is that changes in variance are correlated with changes in the mean level, and the latter are captured by the model.

An important part of the proposed method is the choice of distribution. The data shows a clear positive skew and previous works have modeled this type of data using distributions such as Weibull or Rayleigh \cite{muraleedharan2007modified}. However, the most appropriate distribution depends on the input data.

\section{Conclusions}\label{sec:conclusions}

This paper tackles SWH forecasting problems, which is a relevant topic in ocean data analytics. We are particularly interested in predicting extreme values of SWH, and frame this task as an exceedance probability estimation problem.

We proposed a new method to estimate exceedance probability using the point forecasts produced by an auto-regressive forecasting model. This approach is an alternative to using a binary classification model or an ensemble-based approach. The proposed solution involves transforming the point forecasts into exceedance probabilities based on the CDF. In this particular case study, we resort to the Weibull distribution, but the best distribution depends on the input data.

The experiments carried out suggest that the proposed approach, coupled with a forecasting model based on deep learning, leads to better performance than a binary probabilistic classifier or an ensemble of forecasts. In future work, we will generalize the framework to other application domains besides SWH forecasting.

\section*{Acknowledgments}
The work of L. Torgo was undertaken, in part, thanks to funding from the Canada Research Chairs program.
We regret to inform that L. Torgo passed away during the development of this article. We would like to express our condolences to their family and colleagues.

\bibliographystyle{spmpsci.bst} 

\end{document}